\documentclass[10pt,twocolumn,letterpaper]{article}

\usepackage[pagenumbers]{cvpr} %

\usepackage[dvipsnames]{xcolor}
\usepackage{microtype}
\usepackage{graphics}

\usepackage{float}

\definecolor{cvprblue}{rgb}{0.21,0.49,0.74}
\usepackage[pagebackref,breaklinks,colorlinks,citecolor=cvprblue]{hyperref}

\title{
KFC-W: Generating 3D-Consistent Videos from Unposed Internet Photos
}
\author{
    Gene Chou\textsuperscript{1} \:\: Kai Zhang\textsuperscript{2}\:\: Sai Bi\textsuperscript{2}\:\: Hao Tan\textsuperscript{2}\:\:
    Zexiang Xu\textsuperscript{2}\\[2pt] Fujun Luan\textsuperscript{2}\:\: 
    Bharath Hariharan\textsuperscript{1}\:\: Noah Snavely\textsuperscript{1} \\
    \\
    \textsuperscript{1}Cornell University \,
    \textsuperscript{2}Adobe Research
}

\begin{document}

\twocolumn[{
\maketitle
\begin{center}
    \begin{minipage}{\textwidth}
        \centering
        \includegraphics[width=\textwidth]{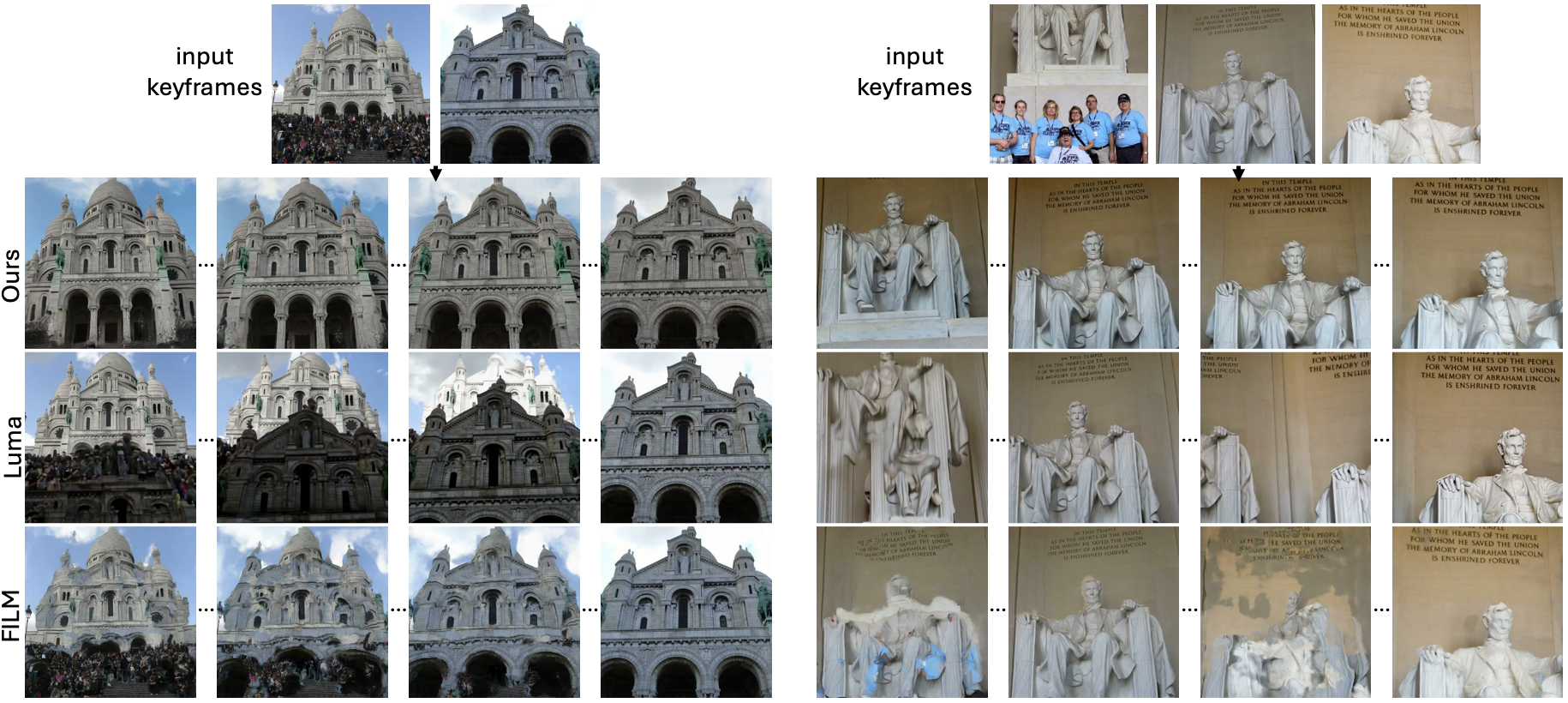}
        \captionof{figure}{Given $n$ unposed input keyframes, the goal is to generate a video of the scene with a realistic camera trajectory and consistent geometry. From top to bottom: Ours, Luma Dream Machine~\cite{luma} (a commercial video generation model), FILM~\cite{reda2022film} (a frame interpolation method). Luma hallucinates new buildings (left scene) and statues (right scene) without understanding the scene layout. FILM is unable to handle wide-baseline inputs and produces blurry transitions. 
        See our \href{https://genechou.com/kfcw}{website} for video playback.
        }
        \label{fig:teaser}
    \end{minipage}
\end{center}
}]

\let\thefootnote\relax\footnotetext{* This work was conducted during Gene's internship at Adobe.}

\begin{abstract}

\vspace{-1em}
We address the problem of generating videos from unposed internet photos.
A handful of input images serve as keyframes, and our model interpolates between them to simulate a path moving between the cameras. 
Given random images, a model’s ability to capture underlying geometry, recognize scene identity, and relate frames in terms of camera position and orientation reflects a fundamental understanding of 3D structure and scene layout. However, existing video models such as Luma Dream Machine fail at this task. 
We design a self-supervised method that takes advantage of the consistency of videos and variability of multiview internet photos to train a scalable, 3D-aware video model without any 3D annotations such as camera parameters.
We validate that our method outperforms all baselines in terms of geometric and appearance consistency. 
We also show our model benefits applications that enable camera control, such as 3D Gaussian Splatting. 
Our results suggest that we can scale up scene-level 3D learning using only 2D data such as videos and multiview internet photos.

\end{abstract}

\vspace{-1em}
\section{Introduction}
\label{sec:intro}

Recent advances in video foundation models~\cite{blattmann2023stable, hong2022cogvideo, yang2024cogvideox, Girdhar2023EmuVideo, jin2024pyramidal} 
learn rich spatio-temporal representations that capture the underlying structure and dynamics of the visual world. It is not surprising that these models contain strong 3D priors that can be used for a variety of downstream applications through finetuning, such as 3D object generation~\cite{han2024vfusion3d, chen2024v3d} and novel view synthesis~\cite{yu2024viewcrafter, kwak2024vivid}. 

In this paper, we further investigate the capabilities of video models to understand 3D structure of real-world scenes. To this end, we propose the task of generating videos from a handful (2-5) of unposed internet photos of the same scene. The generated frames should simulate a path moving between the locations of the cameras, from the first to the second, then from the second to the third, and so on. Given random images, a model’s ability to capture underlying geometry, recognize scene identity, and relate frames in terms of camera position and orientation reflects a fundamental understanding of 3D structure and scene layout. 

Interestingly, we find that this task is challenging for existing video models. 
Even when scaled to a commercial level (e.g. Luma Dream Machine~\cite{luma}), we find that these models often perform 
creative morphing effects rather than produce realistic camera motion. 
We show examples in \cref{fig:teaser}. Frame interpolation methods such as FILM~\cite{reda2022film,danier2023ldmvfi,kalluri2023flavr} are unable to handle wide baselines because they are generally trained on videos with very little camera motion, leading to blurry transitions. 
Luma Dream Machine 
generates high-resolution, sharp videos, but ignores the identity of the scene and creates new structures to fit the input keyframes. In the scene on the left (Sacre Coeur), for instance, Luma transforms the crowd and stairs into a new building. In the scene on the right (Lincoln Memorial Statue), it creates a second statue.

Our main insight is that simply training for general video synthesis is not enough: we need to introduce scalable, 3D-aware objectives. Finetuning with camera poses is an option~\cite{liu2023zero1to3, watson2024controlling}, but collecting 3D annotations is costly~\cite{tung2024megascenes, objaverseXL}. On the other hand, unstructured images and videos abound~\cite{schuhmann2022laion, youtube}.

Thus, we design two objectives. The first is multiview inpainting, which learns 3D priors without 3D annotations. 
The model takes a variable number of condition images, captured from random and wide-baseline viewpoints of a scene, and inpaints an 80\% masked target. Through this process, it learns to extract structural information and scene identity from the condition images, and the illumination and scene layout from the remaining 20\% of the target, in order to fill in the target image accurately. 
Multiview inpainting allows us to leverage the vast corpora of images, which, compared to existing video datasets, provides a greater variety of scenes, more diverse camera viewpoints, and is more accessible~\cite{tung2024megascenes}.
We train using internet photos to adapt to in-the-wild settings, but this objective can be trivially extended to other data sources, such as video, synthetic data, and self-driving datasets for further scaling. 

Our second objective is view interpolation, which takes the start and end frames of a video, and generates intermediate frames.
It requires no annotations from the videos. 

\begin{figure*}[t!]
    \centering 
    \includegraphics[width=\textwidth]{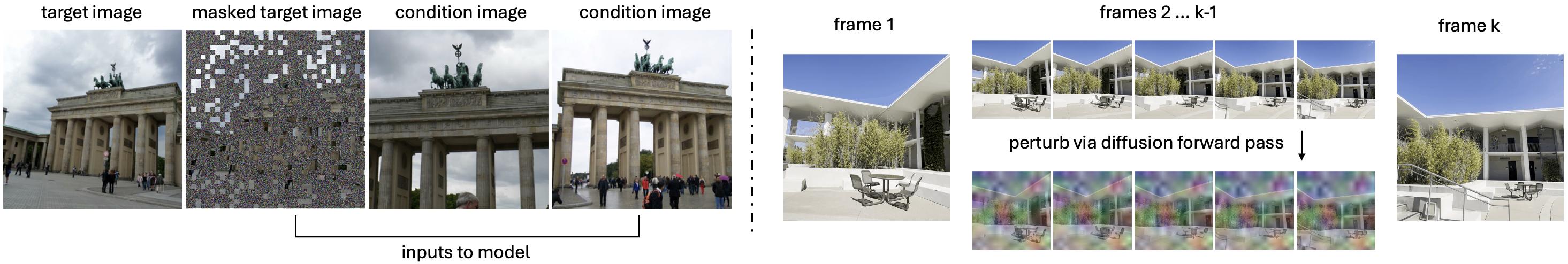}
    \caption{\textbf{Training objectives.} Left: Multiview inpainting. We provide $n$ condition images and one target image to a diffusion model. We add noise to 80\% of the target following the diffusion process. The condition images and remaining 20\% of the target are kept clean. Note how some regions in the target are not seen in the conditions. The model learns priors such as symmetry to generate a plausible image.\\Right: View interpolation. We take $k$ images from a video sequence and add noise to frame 2 to $k-1$ following the diffusion process. The model generates a sequence following a plausible camera path connecting the first and last frames. 
    }
    \label{fig:method-highlevel}
    \vspace{-0.5em}
\end{figure*}

We illustrate these two objectives in \cref{fig:method-highlevel}. These objectives are complementary in enabling our proposed task. Multiview inpainting addresses geometric understanding by training the model to extract 3D relationships from wide-baseline, unposed images. 
View interpolation addresses temporal coherence by training the model to generate smooth, consistent camera trajectories, which is our desired output. 

Finally, we unify these two objectives under the same diffusion denoising objective.
We finetune from a video diffusion model, which adds noise to and denoises selected patches of images: the masked pixels in the multiview inpainting objective, and the intermediate frames in the view interpolation objective. 
This allows us to jointly train both objectives without additional strategies such as pretraining or distillation. 
As a result, our model can take noisy internet photos and produce consistent trajectories, even though it has never seen this input-output pairing during training. 

As shown in \cref{fig:teaser}, our model produces a realistic camera path that captures the layout of the scene, even though the keyframes contain illumination variations and occlusions. For best viewing results, 
please visit our \href{https://genechou.com/kfcw/}{website}, 
where we show side-by-side video playback for comparisons. 
Our approach and results suggest that multiview and video datasets can be complementary to each other even when disjoint.

Following recent works that refer to input images that condition videos as ``keyframes"~\cite{yin2023nuwa-xl, artv, wang2024generative, luma}, and with the added challenge that our keyframes are internet photos, we refer to our approach as \textbf{k}ey\textbf{f}rame-\textbf{c}onditioned video generation in-the-wild (KFC-W). 
In \cref{sec:exp}, 
we evaluate our method by 1) user studies that show it outperforms existing state-of-the-art video generation models; 2) downstream applications such as 3D reconstruction that validate its geometric and appearance consistency.

To sum up, our contributions are as follows:
\begin{enumerate}
    \item We enable consistent video generation conditioned on wide-baseline, in-the-wild keyframes.
    \item We propose a scalable self-supervised training scheme that takes advantage of both multiview and video datasets. The resulting model is 3D-aware without requiring 3D supervision such as camera poses.
    \item We evaluate our method on multiple benchmarks and validate its geometric and appearance consistency. The generated videos can be converted into 3D models (e.g. 3DGS)
    for tasks requiring camera control.
\end{enumerate}

\section{Related Work}
\label{sec:relatedwork}

\noindent \textbf{Reconstruction-based view synthesis.}
One way to synthesize videos is through 3D reconstruction and novel view synthesis, with NeRFs~\cite{mildenhall2021nerf} and 3DGS~\cite{kerbl20233d-3dgs} being popular methods from the past few years. However, even as follow-up work has improved rendering quality~\cite{barron2021mipnerf,barron2023zipnerf,verbin2022refnerf}, 
speed~\cite{kilonerf, baking, mueller2022instant}, 
and has loosened the requirements on number of input views~\cite{verbin2022refnerf, wang2021nerfmm, yu2021pixelnerfneuralradiancefields, fan2024instantsplat} and poses~\cite{bian2022nopenerf, fu2023colmapfree, COGS2024, meuleman2023localrf, lunerf}, 
they are mostly confined to carefully curated captures.
A smaller line of work extends these methods to in-the-wild setting~\cite{li2020crowdsampling,martin2021nerf,sun2022neural,zhang2024gaussian,kulhanek2024wildgaussians} by optimizing from internet photos, but they require dense views, usually at least a few dozen, and preprocessing~\cite{schoenberger2016sfm} to obtain camera parameters.
Even then, the resulting methods lack the ability to fill in unseen viewpoints, leading to floaters and artifacts. 

More recently, starting from LRM~\cite{lrm2024}, transformer-based~\cite{vaswani2017transformer} feed-forward methods~\cite{gslrm2024, wang2023pflrm, wei2024meshlrm, xie2024lrmzero, jin2024lvsmlargeviewsynthesis} aim to perform view synthesis and even reconstruction through learned priors. However, these methods require accurate camera poses during training and testing, and are currently limited to the object-level~\cite{objaverseXL} or high-quality video captures~\cite{RealEstate10k}.

\medskip
\noindent \textbf{Generation-based view synthesis.}
Finetuning image foundation models~\cite{ho2020denoising, rombach2021highresolution} to learn 3D geometry has also proven effective~\cite{watson2022novel, liu2023zero1to3, wu2023reconfusion, zhou2023sparsefusion, liu2023syncdreamer}. 
For example, 3DiM~\cite{watson2022novel} and Zero-1-to-3~\cite{liu2023zero1to3} fine-tune diffusion models to generate novel views conditioned on input views and poses, leveraging and enhancing geometric priors that come from large-scale pretraining. 
Follow-up work like ZeroNVS~\cite{zeronvs, tung2024megascenes, seo2024genwarp, cai2022diffdreamer, yu2023wonderjourney} generalizes to full scenes, but jittering, blurriness, and inconsistency across viewpoints are common in their outputs. 
A number of methods \cite{Yu2024PolyOculusNVS, yu2024viewcrafter, kwak2024vivid, poseguideddiffusion, ren2022look} generates image sets or videos to improve consistency, but 3D supervision is still required during training. This limits scalability since camera parameters for in-the-wild data are typically obtained through SfM~\cite{schoenberger2016sfm, Snavely2006PhotoTE}, which can be computationally expensive and unreliable for sparse views.

\medskip
\noindent \textbf{Video generation and 3D learning.}
Video foundation models~\cite{blattmann2023stable, sora, luma, hong2022cogvideo, yang2024cogvideox, polyak2024moviegencastmedia, BarTal2024LumiereAS, Girdhar2023EmuVideo, singer2022makeavideo, ho2022imagen, jin2024pyramidal} have gained traction due to their potential for immersive storytelling.
They also learn spatial and temporal priors useful for various downstream applications~\cite{chen2024v3d, yu2024viewcrafter, kwak2024vivid, chen2024diffusionforcingnexttokenprediction,
zhang2024physdreamer, pmlr-v235-yang24z}.
For instance, VFusion3D~\cite{han2024vfusion3d} leverages the consistency in video models to generate 3D assets. 
4DiM~\cite{watson2024controlling} jointly trains on video and 3D data for 4D reconstruction. 
In the same spirit, our work finetunes video foundation models with 3D-aware objectives and internet photos to scale up 3D learning in-the-wild.

Relevant to our method is frame interpolation~\cite{jain2024video, luma, reda2022film, kalluri2023flavr, danier2023ldmvfi, wang2024generative, Feng:TRF:ECCV2024}. However, these models are trained on frames with little to no camera motion, but focus on moving objects. Later, we show these methods cannot handle wide-baseline views nor produce realistic camera motion.

\section{Method}
\label{sec:method}

\begin{figure*}[ht!]
    \centering 
    \includegraphics[width=\textwidth]{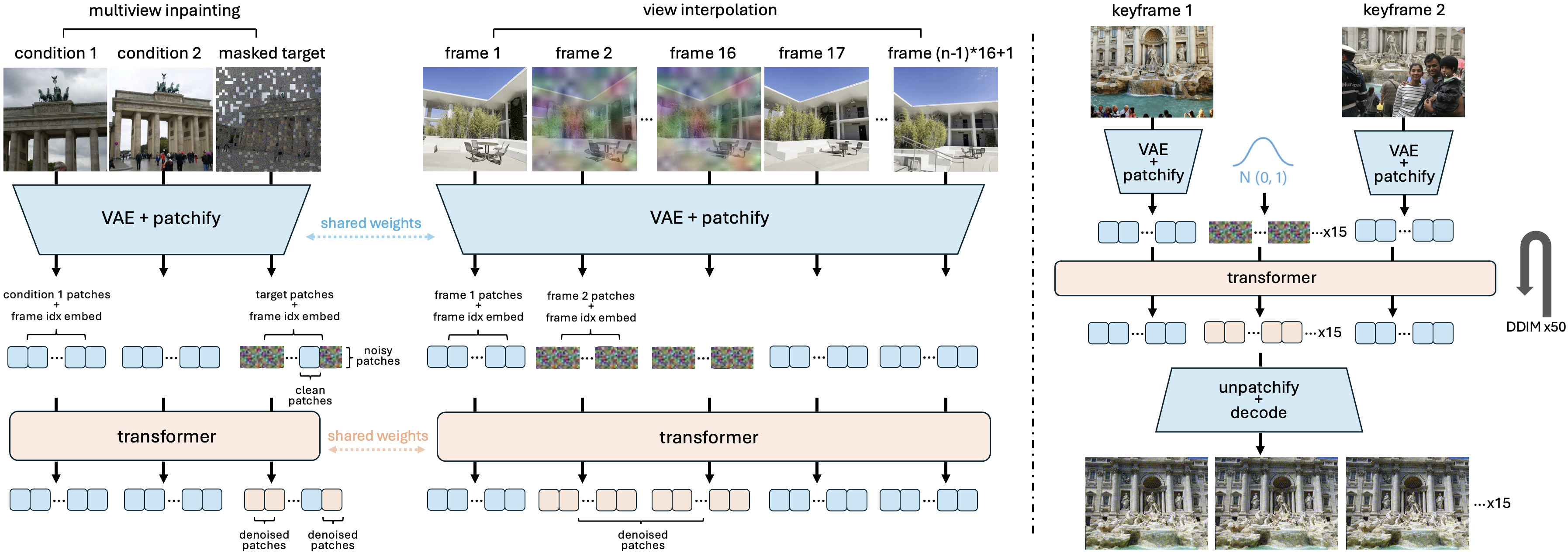}
    \caption{Multiview inpainting of internet photos and view interpolation of videos can be unified under the same denoising objective. \\Left: \textbf{Training.} We denoise the noisy patches (masked patches in multiview inpainting and intermediate frames in view interpolation), while extracting visual information from clean patches (blue patches) via self-attention. Then, we calculate a loss between the denoised (orange) and ground-truth patches. 
    This process operates in latent space.
    \\Right: \textbf{Inference.} Given unposed images of the same scene, we initialize and denoise a fixed number of frames via DDIM. 
    }
    \label{fig:method-arch}
    \vspace{-0.5em}
\end{figure*}

We first provide an overview of keyframe-conditioned video generation in-the-wild (KFC-W). The goal is to generate a sequence of consistent frames given a handful (2-5) of unposed internet photos. 
One core challenge is the absence of supervised training pairs, i.e., internet photos and clean videos from the same scene that we can use as ground truth to resemble our input-output during inference. 
Our approach is we instead take \emph{unpaired} corpora of internet photos and videos, and jointly train two subtasks on each kind of data with the same model: multiview inpainting and view interpolation, shown in \cref{fig:method-highlevel}. 

For multiview inpainting, the only data annotation is that we take images from the same scene. 
Sources such as Wikimedia Commons provide millions of such multiview images with detailed indices and labels. Thus, we use MegaScenes~\cite{tung2024megascenes}, which contains 8M internet photos from 430k scenes labeled by Wikimedia Commons (we do not use any 3D supervision in the dataset). 
Although techniques such as Structure-from-Motion (SfM)~\cite{schoenberger2016sfm} create 3D labels like camera poses, this step is computationally expensive and prone to failure. 
For instance, only 80k, or 1/5, of the scenes in MegaScenes are reconstructed successfully using SfM. 
Our approach bypasses these failure modes and can trivially extend to other sources of data, such as video~\cite{youtube}, synthetic data~\cite{dai2017scannet}, and self-driving datasets~\cite{waymo}.

For view interpolation, we use RealEstate10k~\cite{RealEstate10k} and DL3DV~\cite{ling2023dl3dv}. Both datasets capture sequences of frames within a short time frame, without noticeable changes in illumination. They also do not contain dynamic objects, which are outside the scope of this paper. 

Both objectives are jointly trained on a latent Diffusion Transformer (DiT)~\cite{Peebles2022DiT, rombach2021highresolution}. All images are passed through a pretrained VAE~\cite{vae-kingma} encoder before further processing, including patchifying and adding noise based on diffusion processes. The image patches are then passed through the transformer. This process is shown in \cref{fig:method-arch}.

Next, we explain our self-supervised approach (Sec.~\ref{sec:multiviewinpainting}, \ref{sec:viewinterpolation}) and inference and training details (Sec.~\ref{sec:inference}, \ref{sec:trainingdetails}).

\subsection{Multiview Inpainting}
\label{sec:multiviewinpainting}
We develop a task that 1) learns 3D priors without 3D annotations, and 2) learns from unstructured image collections such as internet photos. 

The model takes a variable number of condition images, captured from random and wide-baseline viewpoints of a scene, and inpaints an 80\% masked target. Through this process, it learns to extract structural information and scene identity from the condition images, and the illumination and scene layout from the remaining 20\% of the target, allowing the model to fill in the target image accurately. In the example in  \cref{fig:method-highlevel}, the target image is taken from an angle rotated counter-clockwise, where some of the structure is not directly observed in the conditions. The model must understand priors such as symmetry to fill in the target. This is important for our final task, as input keyframes are sparse and the model must plausibly fill in content unseen in inputs. 

This approach is inspired by CroCo~\cite{croco, croco_v2}, which observed that cross-view masked image modeling teaches strong 3D priors. Our implementation differs from CroCo in a few ways.
CroCo only operates on two images (i.e. cross-view) because it creates a transformer decoder for each image, which becomes memory-heavy for multiple images. It obtains training pairs from in-the-wild images using camera poses and heuristics. It uses a deterministic MAE~\cite{mae} objective which does not allow for probabilistic generation. In contrast, our method leverages self-attention for all images to take an arbitrary number of inputs, requires minimal annotations on data, and generates diverse outputs.

We implement the patchify and masking operations in latent space~\cite{vae-kingma, rombach2021highresolution} since we use a latent DiT. To all patches of the same image, we add a frame index embedding (since transformers are order-agnostic). For three images, the frame index embeddings would be ($0,1,2$) passed through a linear layer. The patches of the condition images and 80\% of the target are kept clean, while we apply the diffusion forward process to the remaining 20\%. We only calculate the loss between these 20\% of patches with ground truth. We also use a frozen semantic segmentation model on the RGB images to ignore transient objects such as people and vehicles; see our supplement for details.

\subsection{View Interpolation}
\label{sec:viewinterpolation}
This task teaches the model to produce smooth, consistent frames given start and end condition images. 

From a video, we randomly sample $16 \times (n-1) + 1$ sequential frames, where $n$ is the number of condition images and $2 \leq n \leq 5$. 
Every 16th frame is a condition image; i.e. the first frame and 17th frame are condition images, and the 15 intermediate frames in between two conditions are targets that we add noise to using the forward diffusion process. 
To all patches of the same image, we add a frame index embedding. We simply pass the order of the images ($0,1,2,..., (n-1)*16$) through a linear layer. We experimented with normalizing frame indices as well as learnable positional encodings, but found no difference.
This simple objective alone teaches the model to interpolate between input images, though as we show in \cref{sec:exp} as an ablation, it is not sufficient for wide-baseline and in-the-wild images. 

One important aspect of generating consistent in-the-wild videos is controlling illumination.
During training, we condition every image on its own CLIP embedding~\cite{pmlr-v139-radford21a-clip}. We pass all images through a frozen CLIP encoder, and reshape the global feature map to the same shape as the image patches (recall these image patches come from encoding the RGB images through a VAE, then patchifying). Then, we simply perform addition before passing the patches into the transformer. We also experimented with modulation~\cite{perez2018film} but did not observe any differences. During inference, we condition all initialized noisy frames on a CLIP embedding of an image with a desired illumination. 

\begin{figure}[t!]
    \centering 
    \includegraphics[width=0.46\textwidth]{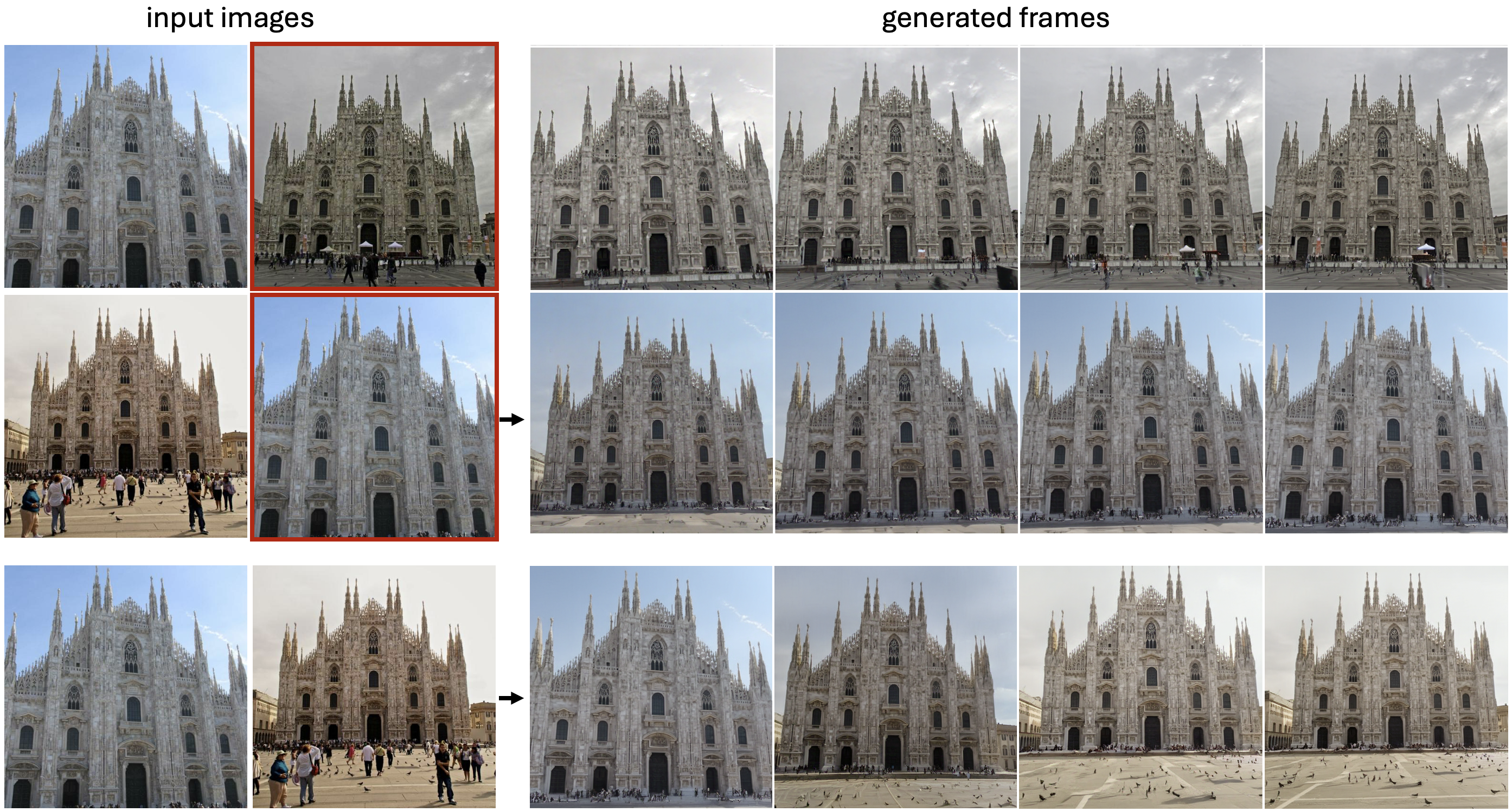}
    \caption{Top two rows: We control illumination by conditioning on the CLIP embedding of the red-bordered image during inference.\\
    Bottom: Without this condition, illumination varies across frames.
    }
    \label{fig:illumination}
    \vspace{-0.5em}
\end{figure}

However, this alone does not force the model to use these embeddings, because the model can simply extract the illumination of the condition images, which, in videos, would be roughly the same as the targets'. Thus, we apply extreme color jittering to the condition images using PyTorch's \verb|ColorJitter| transformation (details in supplement). The model must then extract illumination features from the CLIP embeddings to denoise the intermediate frames, since there is no other source that provides this information. Shown in \cref{fig:illumination}, this mechanism allows us to specify a desired illumination during inference. In the top two rows, we randomly select one image (red-bordered), and condition all intermediate frames on its CLIP embedding. 
The output frames reflect the illumination of the red-bordered image, and remain consistent throughout the sequence. On the other hand, when we do not condition on CLIP embeddings (bottom row), generated frames do not maintain a consistent appearance.

It is worth noting that CLIP embeddings likely only contain coarse information, since its training captions contain terms like ``cloudy," ``sunny," rather than physical properties such as sun angles. Thus, even though we show that this method is capable of controlling coarse illumination, such as the general color scale, there are many possibilities for future work for fine-grained control. 

Finally, we simulate segmentation by randomly masking regions of the condition images, since our input keyframes during inference will also be segmented for transient objects.
See our supplement for details.

\subsection{Inference}
\label{sec:inference}

We show our inference pipeline on the right side of \cref{fig:method-arch}. It combines the two training objectives, with internet photos as input conditions (keyframes), and intermediate video frames as output.
More formally, our input is a set of unposed images of the same scene $\textbf{x} = \{x_1,...,x_n\}$ where $2 \leq n \leq 5$. Our model generates 15 frames in between each input pair $(x_i, x_{i+1})$. We concatenate the frames into a video sequence $\textbf{y}= \{y_1, y_2,...,y_k\}$ where $\{y_1...y_{15}\}$ represents a video path between $x_1$ and $x_2$, and $k=15 \times (n-1)$. 
Thus, the order of the inputs affects the video sequence. To the inputs, we perform segmentation to ignore transient objects such as people and vehicles. To the intermediate frames, we condition them on a CLIP embedding. If not otherwise specified, we use the embedding of the first input.

We run 50 DDIM~\cite{song2020denoising} steps and decode only the intermediate frames. When showing generated videos, we do not include input keyframes as they may contain occlusions. 

\subsection{Training Details}
\label{sec:trainingdetails}
We use a latent Diffusion Transformer (DiT)~\cite{Peebles2022DiT, rombach2021highresolution}.
When training on view interpolation we input sequences of 17, 33, 49, or 65 frames, corresponding to 2, 3, 4, and 5 keyframes, respectively. Our approach extends to longer videos or denser interpolated frames %
at the cost of compute (i.e., transformer sequence length). 
We chose 15 intermediate frames as a balance between compute and visual quality.

An interesting observation is the model's ability to infer which task to perform based on the format of the input. We do not provide flags for context switching, but
simply pass in a number of images corresponding to the objective. Intuitively, both objectives are sufficiently similar to the model, as condition images convey scene identity, and CLIP embeddings supply illumination information. The only difference is the output format and length which can be determined based on the frame indices.

We finetune an internal pretrained text-to-video model, though our approach can be applied to any text-to-image or text-to-video model based on a transformer architecture. We inject our extra condition information -- frame index embedding and CLIP embedding -- by simply adding them to image patch tokens. 
Observations suggest that the specific model and checkpoint we use is comparable to state-of-the-art open-source video models such as CogVideoX-5B~\cite{yang2024cogvideox}. 
We train our model using MegaScenes, Re10k, and DL3DV on 32 NVIDIA A100 80G GPUs for 3 days. 
In the supplement, we provide details for reproducing our method.

\section{Experiments and Evaluation}
\label{sec:exp}

In the following experiments, 
we test on the Phototourism dataset~\cite{Jin2020imcpt} from the Image Matching Challenge, which contains images of 21 landmarks around the world collected from the internet, as well as the official test split of RealEstate10k (Re10k)~\cite{RealEstate10k}.

From each scene in the Phototourism dataset, we randomly sample 3 sets each of 2, 3, 4, and 5 views, leading to 12 sets of sparse views per scene, and 252 sets in total. We call this our \emph{Phototourism} test set. 
From the first 50 scenes from the Re10k test set, we randomly sample 3 sets each of 2, 3, 4, and 5 views, leading to 12 sets of sparse views per scene, and 600 sets in total. We call this our \emph{Re10k} test set. 

\begin{figure}[t!]
    \centering 
    \includegraphics[width=0.45\textwidth]{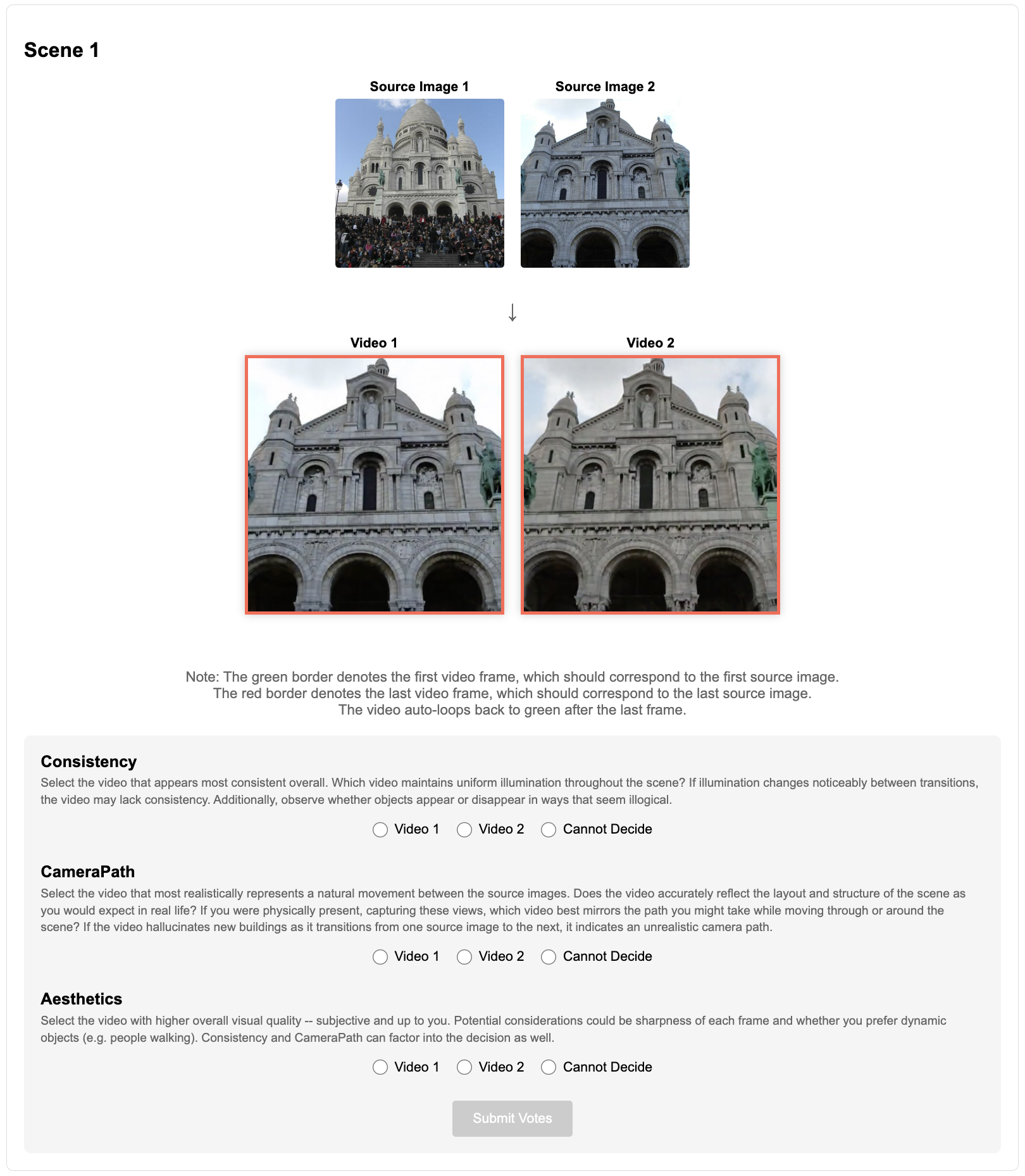}
    \caption{Example scene from our user study interface. We provided detailed descriptions for three criteria: Consistency, CameraPath, and Aesthetics. For each scene, users are asked to express a preference between our results and those of a random baseline.
    }
    \label{fig:userstudy}
    \vspace{-0.5em}
\end{figure}

We perform the following evaluations:

1. We run video generation conditioned on the testing images across multiple baselines, and conduct a user study that ask users to express a preference between pairs of results according to three separate criteria: ``Consistency," ``Camera path," and ``Aesthetics." We show the user study interface and detailed descriptions of each criterion in \cref{fig:userstudy}.

2. We validate the consistency of our generated frames in 3D geometry and appearance using two downstream applications. 
First, we run COLMAP~\cite{schoenberger2016sfm} on the original sparse views, then include our generated frames. This experiment tests whether the generated frames are geometrically consistent with the original views and provide support for feature correspondences.
Second, we optimize a 3D scene with 3D Gaussian Splatting (3DGS)~\cite{kerbl20233d-3dgs} on the original sparse views, then on our generated frames. 
3DGS minimizes a rendering-based reconstruction loss and requires input images be consistent in appearance.

\begin{table}[t]
    \caption{User study results. For a given scene, users vote between ours and a baseline. Our method outperforms commercial models such as Luma on all three criteria.}
    \vspace{-0.5em}
    \centering
    \resizebox{0.9\linewidth}{!} {%
      \begin{tabular}{lccc}
    \toprule
     & \multicolumn{3}{c}{Win Rate of Ours (Full) on Phototourism} \\
    \cmidrule(lr){2-4}
    vs. & Consistency & Camera Motion & Aesthetics \\
    \midrule 
    FLAVR~\cite{kalluri2023flavr} & 100.0\% & 100.0\% & 100.0\% \\
    LDMVFI~\cite{danier2023ldmvfi} & 100.0\% & 100.0\% & 100.0\% \\
    FILM~\cite{reda2022film} & 100.0\% & 100.0\% & 100.0\% \\
    Ours (Video-Only) & 100\% & 96.67\% & 96.67\% \\
    Luma~\cite{luma} & 60.21\% & 73.63\% & 60.21\% \\
    \bottomrule
    \end{tabular} %
    }
    \vspace{1em}
    \resizebox{0.9\linewidth}{!} {%
      \begin{tabular}{lccc}
    \toprule
     & \multicolumn{3}{c}{Win Rate of Ours (Full) on Re10k} \\
    \cmidrule(lr){2-4}
    vs. & Consistency & Camera Motion & Aesthetics \\
    \midrule 
    FLAVR~\cite{kalluri2023flavr} & 100.0\% & 100.0\% & 100.0\% \\ 
    LDMVFI~\cite{danier2023ldmvfi} & 100.0\% & 100.0\% & 100.0\% \\
    FILM~\cite{reda2022film} & 99.50\% & 81.00\% & 99.50\% \\
    Ours (Video-Only) & 76.55\% & 71.31\% & 77.02\% \\
    Luma~\cite{luma} & 83.65\% & 84.86\% & 68.27\% \\
    \bottomrule
    \end{tabular} %
    }
    \label{tab:videocomp}
    \vspace{-0.5em}
\end{table}

\begin{figure}[t!]
    \centering 
    \includegraphics[width=0.46\textwidth]{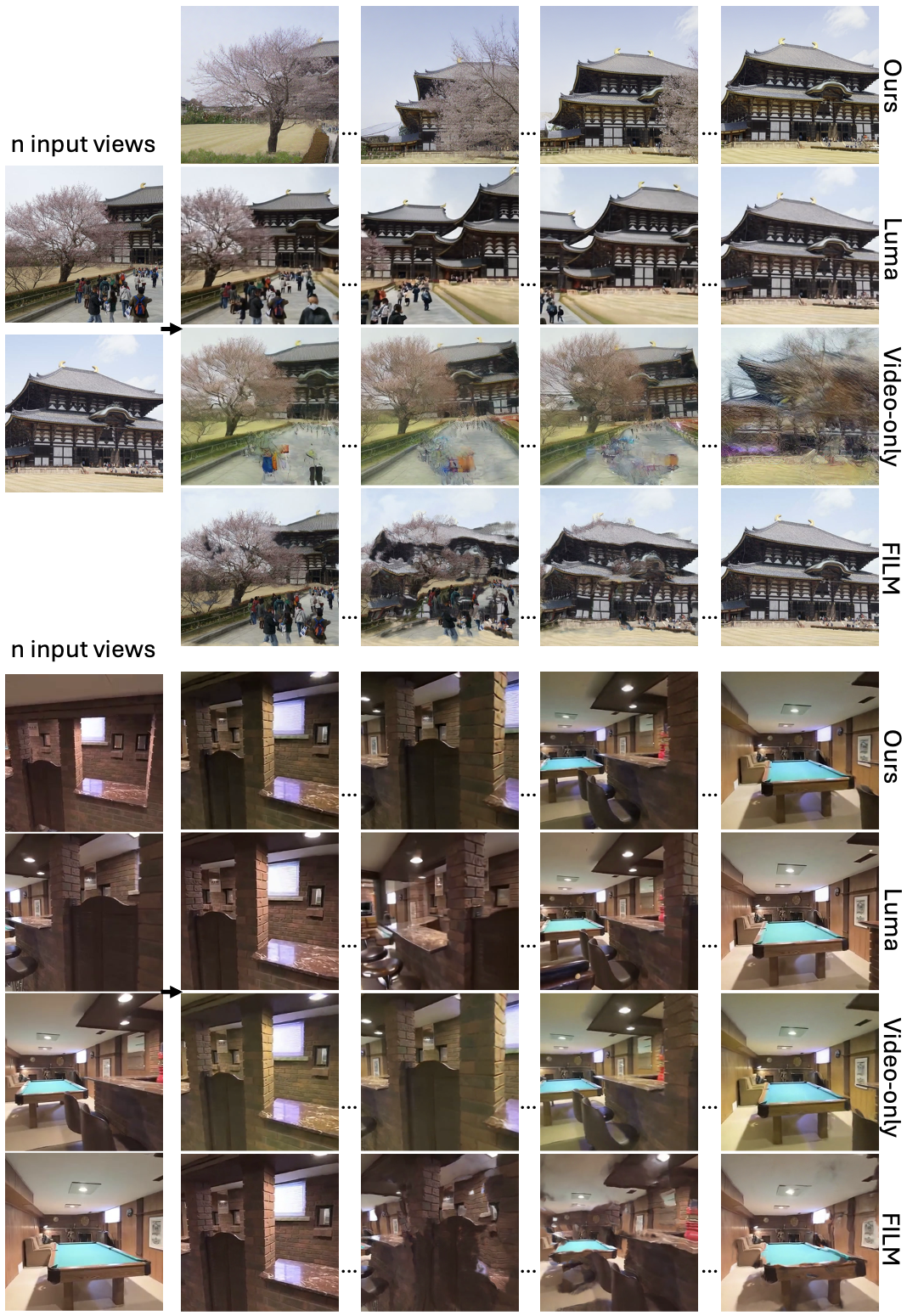}
    \caption{From top to bottom: Ours (Full), Luma, Ours (Video-only), FILM.
Luma hallucinates new buildings (top scene) and produces jittering motions (bottom). Ours (Video-only) generates consistent videos on Re10k but not Phototourism. FILM is unable to handle wide-baseline inputs and produces blurry transitions.
    }
    \label{fig:results2}
    \vspace{-0.5em}
\end{figure}

\subsection{Video Interpolation and Generation}
\label{sec:userstudy}
\noindent \textbf{Baselines.}
We compare to the following methods: FILM~\cite{reda2022film}, FLAVR~\cite{kalluri2023flavr}, LDMVFI~\cite{danier2023ldmvfi}, and Luma Dream Machine~\cite{luma}. The first three are open-source frame interpolation methods, while Luma is a commercial video generation model and we use its paid Dream Machine API. We note that, to the best of our understanding, Luma is the only publicly accessible large-scale model that is capable of interpolating between wide-baseline views. Stable Video Diffusion~\cite{blattmann2023stable}, Pika Labs~\cite{pikalabs}, Emu Video~\cite{Girdhar2023EmuVideo}, and CogVideoX~\cite{yang2024cogvideox} do not support this feature. Runway's Gen-3 Alpha~\cite{runway_gen3_alpha} supports ``frame-interpolation," where input images must have little to no camera motion. We were not able to produce any reasonable results with it on our task.

For methods that can only take two input images, we simply concatenate the videos generated from sequential pairs of keyframes.

We also add an additional baseline: Ours (video-only). This is our method without the multiview inpainting training objective, trained solely on the view interpolation objective using DL3DV and Re10k, with all other training details kept identical. 
This ablation allows us to understand the effect of the multiview inpainting objective. It also serves as a comparison to a second large-scale video generation model (apart from Luma), though finetuning is necessary because our internal model was pretrained with text conditions only. We denote our full method Ours (Full).

\medskip
\noindent \textbf{User study setup.}
For our study, we randomly sampled 25 scenes: 15 from the Phototourism dataset, and 10 from Re10k. 
Our method is compared to each baseline via pairwise comparisons. Users are shown two videos generated from the same input frames and are asked to select which method they prefer according to each evaluation criterion (or the user can select ``Cannot Decide''). 
We show the user study interface in \cref{fig:userstudy}, which also includes detailed descriptions of each criterion. 
When tallying the results, a direct vote counts as 1 point, and a ``Cannot Decide" option counts as 0.5 votes each. For instance, two identical videos should both get 0.5 votes on each criterion, leading to a 50\% preference rate for that specific matchup.
We calculate how often our method is preferred over each baseline, shown in \cref{tab:videocomp}, with results on the two datasets shown separately. 

\medskip
\noindent \textbf{Comparing to frame interpolation methods.}
For comparisons to frame interpolation baselines (FILM, FLAVR, LDMVFI), we collected responses from 10 users, with a total of 150 votes on the Phototourism test set, and 100 votes on the Re10k test set.

As shown in 
\cref{tab:videocomp}, 
none of the baselines produce competitive video sequences on the Phototourism dataset. Their training data consists of small baselines or fixed cameras, and are trained to model dynamics rather than camera motion. 
On the Re10k dataset, only FILM can produce logical generated frames (and only when the camera motion is small), leading to a few ``Cannot Decide" votes. 
See \cref{fig:results2} (bottom scene is from Re10k) and our 
\href{https://genechou.com/kfcw}{website} for examples. .

\medskip
\noindent \textbf{Comparing to our ablated model and Luma.}
For comparisons to Ours (Video-only) and Luma, we collected reponses from 42 users, with a total of 587 votes on the Phototourism test set, and 340 votes on the Re10k test set. 

Our video-only baseline (i.e., our model trained only with view interpolation) works well on the Re10k test set, as expected. 
Images from Re10k scenes are generally closer together, and the model has seen imagery from this domain during training. 
Many videos are nearly identical to Ours (Full), leading to a number of ``Cannot Decide" ratings, although viewpoints with greater distances still led to flickering. 
On the other hand, this method cannot handle the wide-baselines and variability that appear in the Phototourism dataset, even though we used augmentation techniques such as color jittering and random masking during training. 
This demonstrates that introducing internet photos into training is crucial for dealing with in-the-wild scenes. 

For Luma, their Dream Machine API allows users to specify a prompt as well as a start and end keyframe. We set the prompt to ``consistent illumination and smooth camera path" for all videos. 
Luma, as a commercial model, is sampled more densely and at a higher resolution, and trained on substantially more data on a larger model than ours. 
However, our method outperforms Luma on all metrics. 
Luma struggles in several ways on this task: there are noticeable illumination changes as Luma transitions from one source image to the next (``Consistency"). 
Luma also tends to hallucinate new buildings and structures (``Camera path"), shown in \cref{fig:teaser} and \cref{fig:results2}. 
This indicates that Luma does not understand the layout of the scene, and is instead generating morphing effects to match the condition images. 
We believe that Luma achieves a lower preference rate on ``Aesthetics" due in part to these artifacts. In our own observations, Luma's frames can sometimes appear sharper, and dynamic objects such as people walking are modeled more realistically. This led Luma to be preferred for ``Aesthetics" in many (but not most) comparisons.

Surprisingly, Luma was rated as worse on Re10k scenes than on the Phototourism scenes, according to how often our method is preferred. 
Upon observation, we saw that its artifacts tend to be more noticable in indoor scenes, marked by issues such as disappearing objects. 
Furthermore, Luma produces random jittering motions when the input viewpoints are very nearby, possibly to fill time.
Please refer to our 
website for side-by-side comparisons of generated videos. 

These observations show our multiview inpainting objective teaches our model to be 3D-aware and understand depth and symmetry, leading to more realistic video sequences.

\begin{figure}[t!]
    \centering 
    \includegraphics[width=0.45\textwidth]{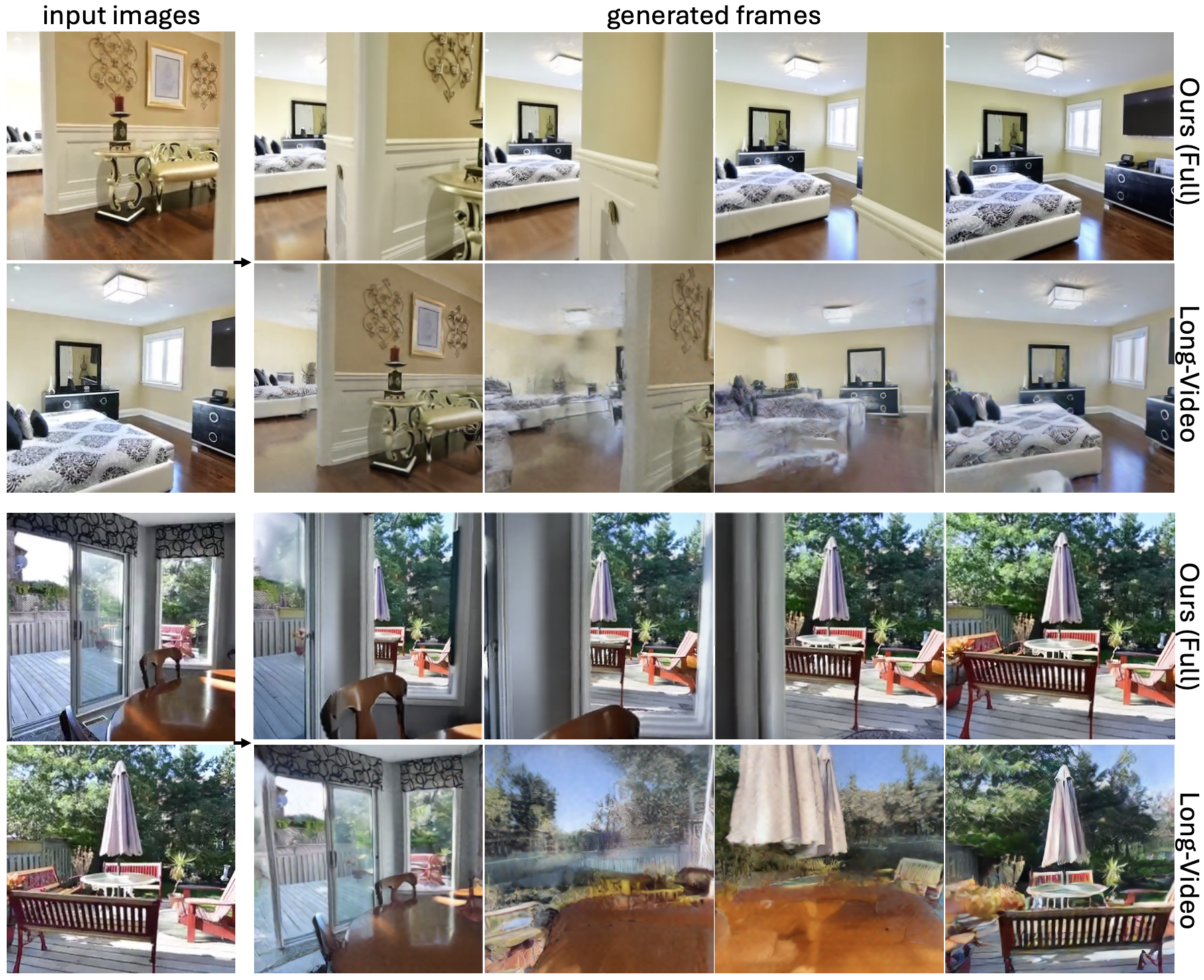}
    \caption{We run an ablation ``Long-Video" with only the view-interpolation objective, but trained on inputs with up to 5x wider baselines. It fails to generalize to inputs with minimal overlap, while Ours (Full) still understands the scene layout.
    }
    \label{fig:longvideo}
\end{figure}

\medskip
\noindent \textbf{Multiview inpainting improves general view interpolation.}
Since Ours (Video-only) works well on Re10k, we verify whether training only on videos, but with wide-baseline inputs, can replace multiview inpainting and internet photos. Thus, we run an ablation similar to Ours (Video-only), but with up to 5x wider baselines as conditions during training. Shown in \cref{fig:longvideo}, this model (``Long-Video") fails on inputs with minimal overlap, while Ours (Full) still understands the scene layout. 
This shows multiview inpainting combined with internet photos improves general view interpolation, not just on in-the-wild data.
We argue that internet photos contain diverse viewpoints, such as extreme rotations and zooming, that cannot be easily learned from videos.

\begin{table}[t]
    \centering
    \caption{Comparison of COLMAP reconstructions when run on original sets of sparse internet photos vs.\ adding in generated frames. Ours (Full) generates geometrically consistent images that provide feature correspondences.
    }
    \vspace{-0.5em}
    \resizebox{\columnwidth}{!}{
        \begin{tabular}{lcc}
            \toprule
            COLMAP SfM & Success Rate & Registered Images \\
            \midrule
            Only Internet Photos & 115/252 & 378/882  \\
            + Generated views (Full) & \textbf{235}/252 & \textbf{741}/882\\
            + Generated views (Video-only) & 179/252 & 589/882\\
            \bottomrule
        \end{tabular}
    }
    \label{tab:sfm-comparison}
    \vspace{-0.5em}
\end{table}

\begin{table*}[t]
    \caption{We run InstantSplat on sparse input images vs our generated frames and compare rendering metrics on 10 random images from each scene. Our method produces densely sampled frames while being consistent in illumination and geometry, leading to improved metrics.}
    \vspace{-0.5em}
    \centering
    \resizebox{\linewidth}{!} {%
    \begin{tabular}{l cccccc cccccc}
    \toprule
    & \multicolumn{6}{c}{\textbf{Phototourism}} & \multicolumn{6}{c}{\textbf{Re10k}} \\
    \cmidrule(lr){2-7} \cmidrule(lr){8-13}
    & \multicolumn{3}{c}{On Internet Photos} & \multicolumn{3}{c}{On Generated Frames} & \multicolumn{3}{c}{On Original Frames} & \multicolumn{3}{c}{On Generated Frames} \\
    \cmidrule(lr){2-4} \cmidrule(lr){5-7} \cmidrule(lr){8-10} \cmidrule(lr){11-13}
    Method & PSNR$(\uparrow)$ & SSIM$(\uparrow)$ & LPIPS$(\downarrow)$ & PSNR$(\uparrow)$ & SSIM$(\uparrow)$ & LPIPS$(\downarrow)$ & PSNR$(\uparrow)$ & SSIM$(\uparrow)$ & LPIPS$(\downarrow)$ & PSNR$(\uparrow)$ & SSIM$(\uparrow)$ & LPIPS$(\downarrow)$ \\
    \midrule
    InstantSplat & 11.701 & 0.3510 &  0.5703 & \textbf{13.960} & \textbf{0.4123} &  \textbf{0.4864 }& 19.857 & 0.7269 & 0.2663 & \textbf{21.798} & \textbf{0.7916} & \textbf{0.2190} \\
    \bottomrule
    \end{tabular}
    }
    \label{tab:renderingnvs}
    \vspace{-0.5em}
\end{table*}

\begin{figure}[t!]
    \centering 
    \includegraphics[width=0.45\textwidth]{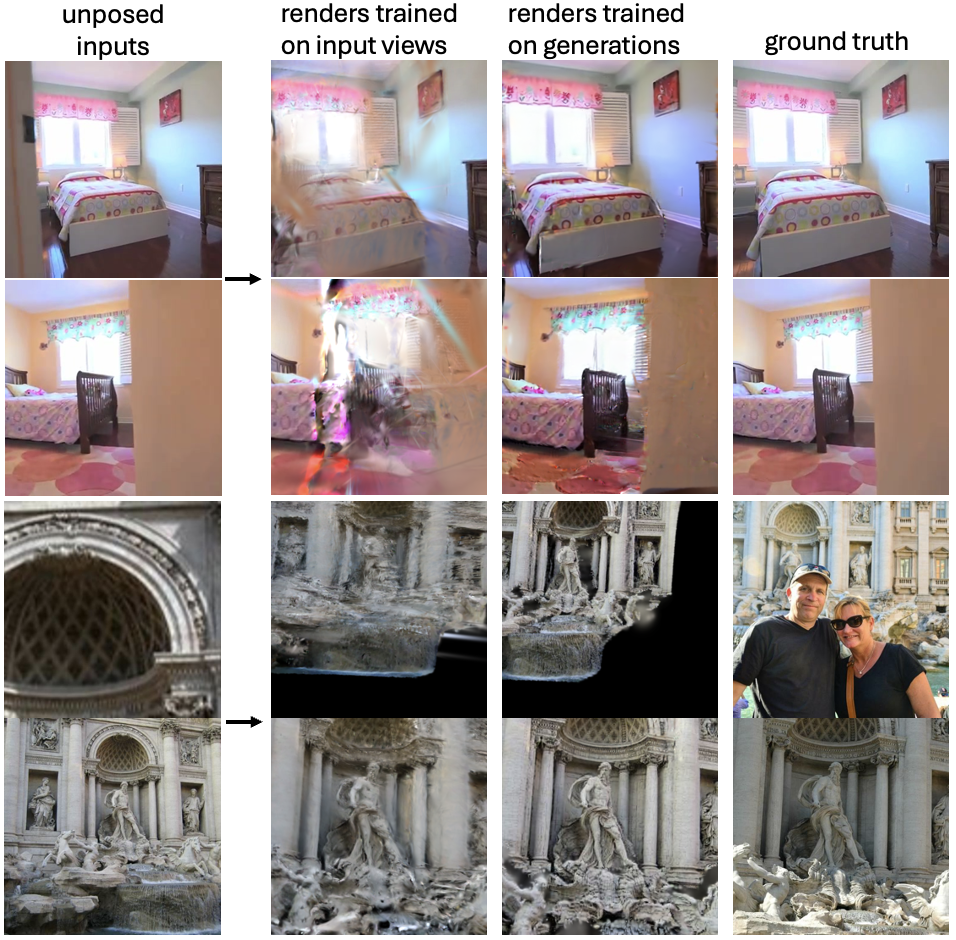}
    \vspace{-0.5em}
    \caption{We run InstantSplat~\cite{fan2024instantsplat} on original input views and on generated frames. The rendered results on generated frames exhibit fewer artifacts and sharper content. 
    }
    \label{fig:instantsplat}
    \vspace{-0.5em}
\end{figure}

\subsection{Application 1: SfM Reconstruction}

We validate whether our generated frames are consistent in geometry, and therefore suitable for downstream applications such as 3D reconstruction.
We run COLMAP~\cite{schoenberger2016sfm} 
on the original input views, then include our generated frames. 
COLMAP is unreliable on sparse views because when viewpoints are too far apart, COLMAP struggles to find sufficient matching features between pairs of images. 
As shown in \cref{tab:sfm-comparison}, only 45\% of sets of sparse views are successfully reconstructed (``Only Internet Photos"). 
Of the 882 images total, 43\% were registered. On the other hand, COLMAP successfully reconstructed 93\% of sets of views when we included our video frames, which were generated from the same sets of sparse views (``+ Generated views (Full)"). 
Now, 84\% of the original frames could be registered, nearly doubling the amount. This shows our generated frames provide reliable feature correspondences that connect distant views.

However, not all view interpolation methods achieve this effect. We also run this experiment by adding generated frames from the ``Ours (Video-Only)" baseline, denoted ``+ Generated views (Video-only)". There was significantly less improvement: 71\% success rate and 67\% registered images.

\subsection{Application 2: 3D Gaussian Splatting}

We experiment with running 3D Gaussian Splatting (3DGS) on our generated frames. 
In contrast to COLMAP, which is robust to illumination changes, the input images for (vanilla) 3DGS must be consistent in appearance and illumination. 
To validate our method, we compare running 3DGS on original inputs vs.\ our generated frames (without the original inputs).
We use InstantSplat~\cite{fan2024instantsplat}, a 3DGS method that builds on COLMAP-Free-3DGS~\cite{fu2023colmapfree} and DUSt3R~\cite{dust3r_cvpr24} to generate Gaussian Splats given sparse, unposed images. 

We show results in \cref{tab:renderingnvs} and \cref{fig:instantsplat}. ``On Internet Photos" and ``On Original Frames" refer to training each scene using the original sparse views in the test sets, while ``On Generated Frames" refer to using our model's output when conditioned on those same sparse views. 

For each scene, we train with the default settings in the InstantSplat open-source code. Then, we sample 10 images not used for training from the scene, register their poses to the same coordinate frame, and render them using the trained model to report PSNR, SSIM~\cite{wang2004ssim}, and LPIPS~\cite{zhang2018perceptual} metrics. We provide details in the supplement. 

Internet photos from the Phototourism dataset have wide baselines, significant occlusions, and varying illumination, which make it very difficult to train 3DGS methods based on a pixel-rendering loss. 
Our generated frames are denser and with more consistent illumination, leading to substantial improvements in reconstruction metrics. 
Even though the Re10K dataset has similar illumination conditions and smaller baselines across frames, we observe an improvement across all metrics as a result of training with denser frames. We provide rendered sequences on the website.

Note that the metrics fo the Phototourism dataset are much lower than those for Re10k because the testing ground-truth is also internet photos, which lead to high rendering losses (see the ground-truth column in \cref{fig:instantsplat}).

Through these applications, we show that video models trained with 3D-aware objectives can be useful as 3D priors for various downstream tasks.

\section{Future Work and Discussion}
\label{sec:conclusion}

\noindent\textbf{Future work.} 
Potential future work include scaling up data to model dynamic objects, enforcing a fine-grained constraint on illumination, and extrapolating to unseen views.

\medskip
\noindent\textbf{Scaling 3D awareness via self-supervised learning.} We posit that brute-force scaling will not help video models understand the physical world, as even the most advanced video models today have difficulty understanding physics~\cite{kang2024how} or scene layouts. 
However, rather than incorporating conditions such as camera poses, which can be difficult to reliably estimate at scale, 
we jointly train a scalable 3D-aware objective. 
We suggest that this concept can be applied to other tasks as well, such as modeling motion that respects physical constraints. 
Additionally, our model generates videos from internet photos even though it never sees this specific input-output pairing during training, suggesting multitask learning leads to emergent capabilities.

\section*{Acknowledgments}
We thank Kalyan Sunkavalli and Nathan Carr for supporting this project. Gene Chou was supported by an NSF graduate fellowship. 

{
    \small
    \bibliographystyle{ieeenat_fullname}
    \bibliography{main}
}

\end{document}